\documentclass[journal]{IEEEtran}
\usepackage{graphicx}
\usepackage{verbatim}
\usepackage{amstext}
\usepackage{color,xcolor}
\usepackage{amsmath}
\begin{document}
%
\title{Tiny Object Tracking: A Large-scale Dataset and A Baseline}
%
%

\author{Yabin Zhu, Chenglong Li, Yao Liu, Xiao Wang, Jin Tang, Bin Luo, \emph{Senior Member, IEEE}, Zhixiang Huang
\thanks{Y. Zhu is with Key Laboratory of Intelligent Computing and Signal Processing of Ministry of Education, Key Laboratory of Electromagnetic Environmental Sensing of Anhui Higher Education  Institutes, Anhui Provincial Key Laboratory of Multimodal Cognitive Computation, Anhui University, Hefei 230601, China. (email: zhuyabin0726@foxmail.com)

C. Li is with Anhui Provincial Key Laboratory of Multimodal Cognitive Computation, School of Artificial Intelligence, Anhui University, Hefei 230601, China. (email: lcl1314@foxmail.com)

Y. Liu, J. Tang and B. Luo are with School of Computer Science and Technology, Anhui University, Hefei 230601, China. (email: liuyao.study@foxmail.com; tangjin@ahu.edu.cn; ahu\_lb@163.com))}
\thanks{Xiao Wang is with Peng Cheng Laboratory, Shenzhen, China. (email: wangxiaocvpr@foxmail.com)

Z. Huang is with Key Laboratory of Intelligent Computing and Signal Processing of Ministry of Education, Key Laboratory of Electromagnetic Environmental Sensing of Anhui Higher Education  Institutes, Anhui University, Hefei 230601, China. (email: zxhuang@ahu.edu.cn)}  
}

%
%

\markboth{IEEE Transactions manuscript, December~2021}%
{Shell \MakeLowercase{\textit{et al.}}: Bare Demo of IEEEtran.cls for IEEE Journals}

\maketitle

\begin{abstract}
Tiny objects, frequently appearing in practical applications, have weak appearance and features, and receive increasing interests in meany vision tasks, such as object detection and segmentation.
To promote the research and development of tiny object tracking, we create a large-scale video dataset, which contains 434 sequences with a total of more than 217K frames. Each frame is carefully annotated with a high-quality bounding box. In data creation, we take 12 challenge attributes into account to cover a broad range of viewpoints and scene complexities, and annotate these attributes for facilitating the attribute-based performance analysis. 
To provide a strong baseline in tiny object tracking, we propose a novel Multilevel Knowledge Distillation Network (MKDNet), which pursues three-level knowledge distillations in a unified framework to effectively enhance the feature representation, discrimination and localization abilities in tracking tiny objects. 
Extensive experiments are performed on the proposed dataset, and the results prove the superiority and effectiveness of MKDNet compared with state-of-the-art methods. The dataset, the algorithm code, and the evaluation code are available at \textcolor{blue}{https://github.com/mmic-lcl/Datasets-and-benchmark-code}. 
\end{abstract}


%
\IEEEpeerreviewmaketitle

\section{Introduction}
\begin{figure}[]
\centering
\includegraphics[width=\columnwidth]{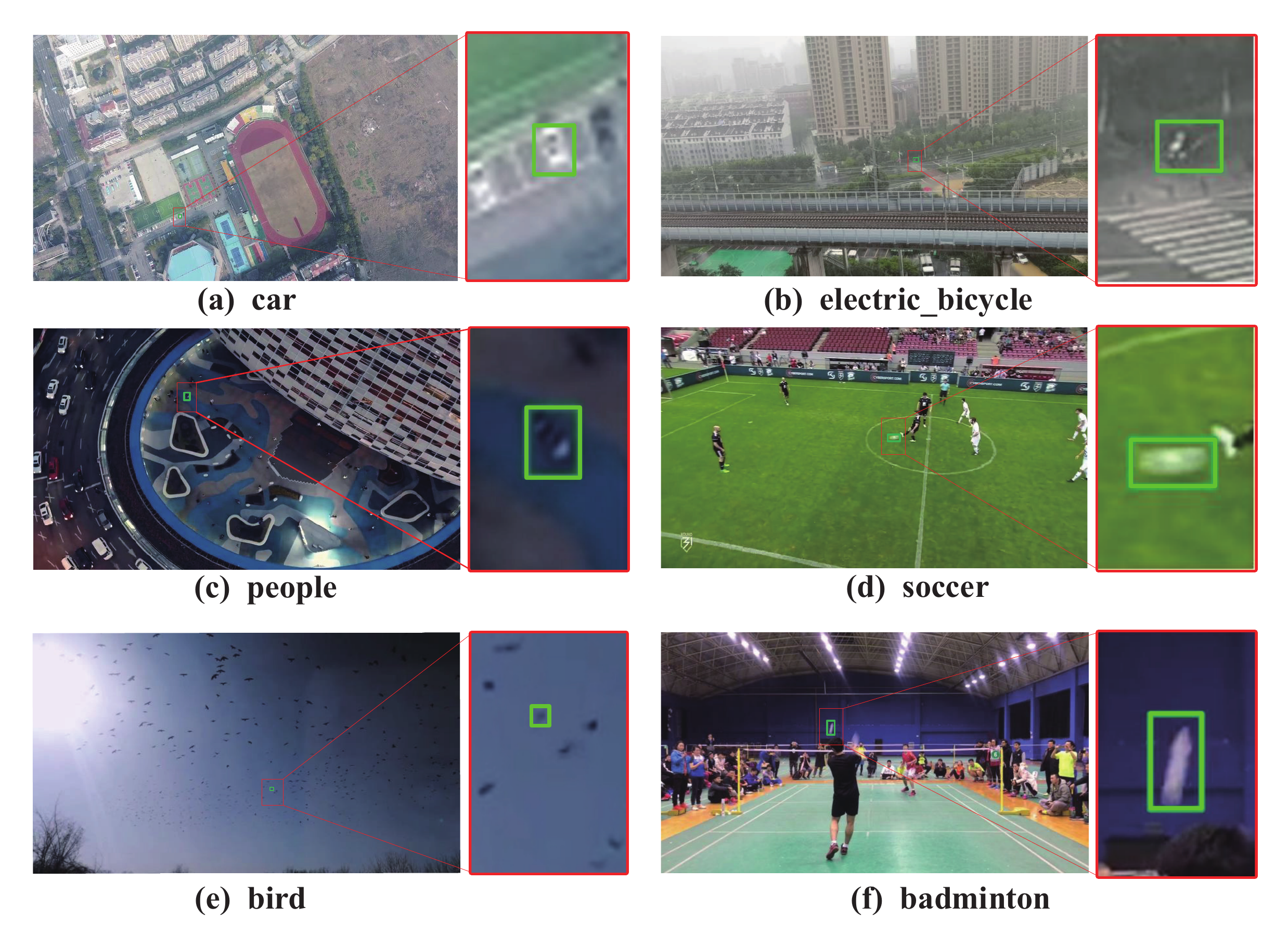} \\
\caption{Some representative samples in the proposed LaTOT dataset.}\label{fig::example}
\end{figure}

Visual tracking ~\cite{yun2018action,2020Antidecay,ge2020cascaded} is an important research topic in the field of computer vision. Although unprecedented progress has been made in recent years, there are still many challenging problems. For example, in some practical applications such as unmanned aerial vehicle, remote sensing and ball game, target objects are usually too small and thus has weak appearance and features, and thus tracking them is very challenging, as shown in~\ref{fig::example}. The perception of tiny and small objects has been concerned in the field of object detection ~\cite{noh2019better,kisantal2019augmentation} and object segmentation~\cite{guo2018small,wang2019miss}, but rare tracking algorithms focus on tiny objects. In this work, we address the problem of tiny object tracking.




In the field of visual tracking, OTB~\cite{wu2013online,Wu2015Object} and VOT ~\cite{2016The,kristan2018sixth,kristan2015visual,kristan2016novel,kristan2019seventh,hadfield2014visual} are two widely used datasets.
However, these datasets are small-scale.
TrackingNet~\cite{muller2018trackingnet},  LaSOT~\cite{fan2019lasot} and GOT-10K ~\cite{huang2019got} are three largest tracking datasets, but they only contain 144, 52 and 42 tiny object videos respectively. It is insufficient for the training and testing of deep trackers on LaTOT.
Small90~\cite{liu2019aggregation} is collected from existing tracking datasets by selecting the small videos whose size ratio of object region is less than 0.01 of the whole image. We argue that such definition sometimes is unreasonable. For example, some trackers~\cite{nam2016learning, bhat2019learning} crop a local region around the target for tracking and their performance is thus not affected by the size ratio. Besides, Small90 also suffers from small-scale size and limited challenges. 

To provide a comprehensive evaluation platform for tiny object tracking, we contribute a large-scale Tiny Object Tracking dataset (LaTOT) which has the following major properties.
\begin{itemize}
\item {\bf Large-scale size}. LaTOT consists of 434 video sequence with a total of more than 217K frames in real-world scenarios. Table~\ref{tb::datasets} shows that the numbers of sequences, categories and total frames of our dataset are far beyond tiny object videos on popular tracking datasets.

\item {\bf Dense and ultra-precision annotation}.
%
For tiny object videos, object regions are usually unclear and accurately annotating them is very hard. Therefore, we enlarge the target region as much as possible and then label each frame accurately, ensuring that the labeling accuracy can reach at least two decimal places. Finally, to avoid labeling errors, the dataset is checked and modified by 16 inspectors and assessors in three rounds.

\item {\bf High diversity and challenging}. LaTOT contains 48 categories for tiny objects capturing from 270 scenes, which are beneficial for the evaluation of different tracking algorithms on real-world scenarios. It contains 12 major attributes which are challenging for current trackers to track. Some representative samples are visualized in Fig.~\ref{fig::example}.

\end{itemize}

As the performance of tiny object tracking is easily affected by low resolution, image blur, and noisy information, etc. We think that all these factors make the responses in feature maps weak, and it is difficult for trackers to discriminate and locate. 
To solve this problem, we propose a novel Multilevel Knowledge Distillation Network (MKDNet) to effectively enhance the feature representation, discrimination and localization abilities for tiny objects. To achieve effective knowledge transfer, we design two networks which share the same structure and initial parameters. The first one is the teacher network and the another one is the student network. Our MKDNet pursues three levels of knowledge distillations, including feature-, score-, and IoU-level distillation. Specifically, the features-level distillation uses the high-resolution features outputted from the teacher network as the supervision to guide the learning of low-resolution tiny object features. The score-level distillation aims to use the potential distribution of classification output from teacher network  as the supervision to guide the learning of student network. 
To improve the localization ability of student network, we also design a IoU-level distillation scheme that uses the IoU score of the teacher as the supervision to guide the learning of student network. 
Since the teacher network sometimes has wrong guidance supervision and direct usage of information from teacher network would lead to misleading learning of student network. Therefore, we define a reliable distillation measure, which takes the loss of the teacher network as the distillation upper bound, to avoid wrong and invalid distillation.

To sum up, our contributions are listed as follows. 
\begin{itemize}
\item We are the first to propose a large-scale dataset for tiny object tracking. It is large-scale size, high diverse and high-quality dense annotations, and thus would greatly promote the research and development of tiny object tracking.

\item We propose a novel baseline method called multi-level knowledge distillation network (MKDNet) to effectively enhance the feature representation, discrimination and localization abilities in tiny object tracking.

\item An adaptive and effective strategy is designed to avoid invalid information propagation in the knowledge distillation phases. A reliability measure is defined as the distillation upper bound based on the loss of teacher network, which enables the learning of student network from teacher network more reliable.

\item Extensive experiments on the proposed dataset validate the superiority and effectiveness of our MKDNet. We also report the results of 24 state-of-the-art trackers on the proposed dataset for the evaluation of different tracking algorithms. We believe our dataset and baseline would promote the research and development of tiny object tracking. 
\end{itemize}

\section{Related Work}

In this section, we give a brief introduction to visual tracking, small object perception, and knowledge distillation. More related works can be found in the following surveys \cite{wang2021knowledge,marvasti2021deep,tong2020recent}.

\subsection{Visual Tracking}
In recent years, a large number of trackers~\cite{nam2016learning,jung2018real,wang2021ganTANetTrack,wang2021dynamic,li2018high,li2019siamrpn++,2019GlobalTrack,voigtlaender2020siam,danelljan2019atom,bhat2019learning} have been proposed to solve various challenges in visual tracking. The first category is trackers based on multi-domain learning, including MDNet~\cite{nam2016learning} and  RT-MDNet~\cite{jung2018real}.
MDNet and RT-MDNet distinguish foreground objects from background online by using the binary classifier.
The second category is Siamese-based trackers, including SiamRPN~\cite{li2018high}, SiamRPN++~\cite{li2019siamrpn++}, GlobalTrack~\cite{2019GlobalTrack} and Siam R-CNN~\cite{voigtlaender2020siam}. SiamRPN and SiamRPN++ add the region proposal network module which not only improves the tracking performance but also the speed. GlobalTrack and Siam R-CNN aim to solve long-term tracking challenge by using wide range of search strategies. The third category is trackers based on discriminant feature learning, including ATOM~\cite{danelljan2019atom} and DiMP ~\cite{bhat2019learning}. ATOM and DiMP are the combination of Siamese network and correlation filtering model in tracking, and they introduce the scale regression method~\cite{jiang2018acquisition} into the tracking framework. 
However, these methods do not focus on the problem of tiny objects, which have many challenges, such as low resolution, image blur, less effective information, and more noises. Therefore, it is difficult for general trackers to track tiny objects accurately.


\subsection{Small Object Perception} 
The perception of small objects lies in multiple tasks such as object detection~\cite{noh2019better,kisantal2019augmentation}, semantic segmentation~\cite{guo2018small}, and object tracking~\cite{marvasti2020comet}. In object detection, J. Noh et al.~\cite{noh2019better} propose a tiny object feature enhancement strategy based on GAN, which effectively alleviates the problem of weak features of small objects. 
M. Kisantal et al.~\cite{kisantal2019augmentation} propose to boost the detection accuracy of small objects by solving the problem of an unbalanced distribution of small objects. In semantic segmentation, D. Guo et al.~\cite{guo2018small} propose a measurement method to effectively increases the contribution of small objects to the overall loss, thereby improving the segmentation accuracy of small target instances. However, these methods can not directly solve the problem of tiny object tracking as it aims to track tiny objects in video sequences instead of images.
In visual tracking, S. Marvasti-Zadeh et al.~\cite{marvasti2020comet} combine the context-awareness strategy and the multi-scale feature aggregation method to solve the object tracking problem under aerial photography. But it is difficult to solve the problem of tiny objects in natural scenes.
But it only aggregates multi-scale features and does not substantially enhance the quality of tiny object. Therefor, it is difficult to solve the problem of tiny objects tracking.

\subsection{Knowledge Distillation} 
A lot of works~\cite{hinton2015distilling, park2021learning, xu2020feature, chung2020feature, saputra2019distilling, kang2021data} use Knowledge distillation to enhance the performance of convolutional neural networks. These methods of knowledge distillation can be roughly divided into three categories, including logits distillation, feature distillation and regression distillation. The logits distillation methods~\cite{hinton2015distilling, park2021learning} mainly enhance the performance of student network by restricting the classification distribution of the student network to be consistent with the teacher network.
The feature distillation methods~\cite{xu2020feature,chung2020feature} are  based on original knowledge distillation. Their core idea is to use not only the output of the teacher network, but also the features of the intermediate layer of teacher network as the knowledge to boost the training of student network. The regression distillation methods~\cite{saputra2019distilling,kang2021data} explore the application of knowledge distillation in the regression model. Inspired by these methods, we propose a  multilevel knowledge distillation network to deal with the problem of tiny object tracking.

\begin{table*}[h]\footnotesize 
\setlength{\belowcaptionskip}{1cm}
\caption{ Comparison of tiny object videos in our LaTOT with popular visual tracking datasets. $"0"$ indicates that there is no data that meets the conditions.
$"-"$ means that there is no statistically relevant data.}
\centering
\renewcommand\arraystretch{1.5}
\begin{tabular}{ c | c c c c c c c c}
\hline
Benchmark& Video &Class &Frames rate &Min frames &Max frames & Avg frames & Avg Duration(seconds) &Total frames \\
\hline
OTB50~\cite{wu2013online} & 0 & 0 & 30 & 0 & 0 & 0 & 0 & 0 \\

OTB100~\cite{Wu2015Object} & 0 & 0 & 30 & 0 & 0 & 0 & 0 & 0 \\

VOT14 ~\cite{hadfield2014visual}& 0 & 0 & 30 & 0 & 0 & 0 & 0 & 0 \\

VOT15~\cite{kristan2015visual} & 2 & 2 & 30 & 41 & 377 & 209 & 7.0 & 0.4K \\

VOT18~\cite{kristan2018sixth} & 4 & 3 & 30 & 41 & 377 & 222 & 7.4 & 0.8K \\

VOT19~\cite{kristan2019seventh} & 6 & 5 & 30 & 41 & 377 & 181 & 6.0 & 1.0K \\

TC128~\cite{liang2015encoding} & 12 & 9 & 30 & 90 & 454 & 193 & 6.4 & 2.3K \\

DTB70~\cite{li2017visual} & 0 & 0 & 30 & 0 & 0 & 0 & 0 & 0 \\

ALOV300~\cite{smeulders2013visual} & 0 & 0 & 30 & 0 & 0 & 0 & 0 & 0 \\

UAV123~\cite{mueller2016benchmark}  & 10 & 6 & 30 & 157 & 577 & 399 & 13.3 & 3.9K \\

UAV20L~\cite{mueller2016benchmark}  & 0 & 0 & 30 & 0 & 0 & 0 & 0 & 0 \\

NfS~\cite{kiani2017need} & 1 & 1 & 240 & 1,785 & 1,785 & 1,785 & 7.43 & 1.7K \\

TNL2K~\cite{wang2021tnl2k} &3 &2 &30 & 368 & 578 & 467 &15.6 &1.4K \\

UAVDT~\cite{yu2020unmanned} & 9 & 7 & 30 & 232 & 838 & 476 & 15.9 & 4.2K\\

TrackingNet~\cite{muller2018trackingnet} & 144 & - & 30 & 130 & 541 & 458 & 15.2& 6.6K \\

LaSOT~\cite{fan2019lasot} & 52 & 17 & 30 & 1,000 & 9,999 & 2887 & 96.2 & 150.K \\

GOT-10k~\cite{huang2019got} & 42 & 17 & 10 & 81 & 601 & 221 & 22.1 & 9.2K \\

Small90~\cite{marvasti2020comet} & 14 & 7 & 30 & 93 & 820 & 347 & 11.6 & 4.8K \\
\hline 
LaTOT &434 &48 &30 &21 &4,632 &501 & 16.7 &217.7K \\\hline
\end{tabular}
\label{tb::datasets}
\end{table*}

\section{LaTOT Dataset} 

In this section, we introduce the details of our large-scale Tiny Object Tracking benchmark  (LaTOT), including definition of tiny object in video, data collection, data annotation, statistical and analysis.

\subsection{Definition of tiny object in video}
In the task of object detection, there are two definitions of tiny object, including \textbf{relative size} and \textbf{absolute size}. 
For the relative size, the area of the target size is smaller than $1\%$ of the whole image. The absolute size denotes that the resolution of target object is less than $32\times32$~\cite{lin2014microsoft}.

In our case, we need to identify tiny objects in video sequences rather than in image. To this end, we define tiny objects in video by considering both relative size and absolute size. In specific, we average the absolute size and relative size of the whole video sequence respectively to obtain the averaging absolute size and the averaging relative size. Then, we determine whether a video is tiny or not by judging that both the average absolute size and the average relative size are below the predefined thresholds. In this work, we set the threshold of averaging relative size as $1\%$, which is same with existing works. Since the size of common tiny objects in videos are about $22\times22$ pixels, and thus we set the threshold of averaging absolute size to $22\times22$.

\subsection{Data Collection and Annotation}
Most sequences of LaTOT are collected from public video platforms, including TikTok, Bilibili, Tencent Video and Xigua Video. There is also a small part of the dataset captured by Huawei nava 7 and Iphone 11. LaTOT contains a total of 434 video sequences with an averaging frames of 501. From Table~\ref{tb::datasets}, the numbers of tiny object sequences, categories and total frames of our dataset are significantly larger than existing popular tracking benchmarks.

Dense and ultra-precision annotation is essential for fair evaluation of trackers. To this end, we manually annotate each video frame in LaTOT, and perform multi-stage careful inspections and modifications. To accomplish the labeling work, we set up an annotation team, which includes a Ph.D. student and 15 master students working in visual tracking field.
Moreover, we choose three team members as the quality assessors. If the annotation results are not unanimously agreed by the quality assessors, they will be sent back to the annotation team for modification.

As shown in Fig.~\ref{fig::example} and Fig.~\ref{fig::Qualitative2}, some target objects are very small and often encounter other challenges such as fast motion, motion blur, low resolution and similar object. It usually results in nearly 60\% of the dataset as unqualified labelled ones in the first round of annotation. To handle this problem, we enlarge each frame for re-annotation to ensure ultra-precision annotations. Moreover, our labeling accuracy can reach two decimal places. After three stages of inspection and labeling by assessors and labellers, the labelling work is painstakingly completed. We do our best to ensure that this dataset has high-quality dense annotations. A \textbf{demo video} of our proposed LaTOT benchmark dataset can be found on this link \footnote{{\textcolor{blue}{https://www.youtube.com/watch?v=IYYTLAOsa-E}}}.

\begin{figure*}[htb]
\centering
\includegraphics[width=\textwidth]{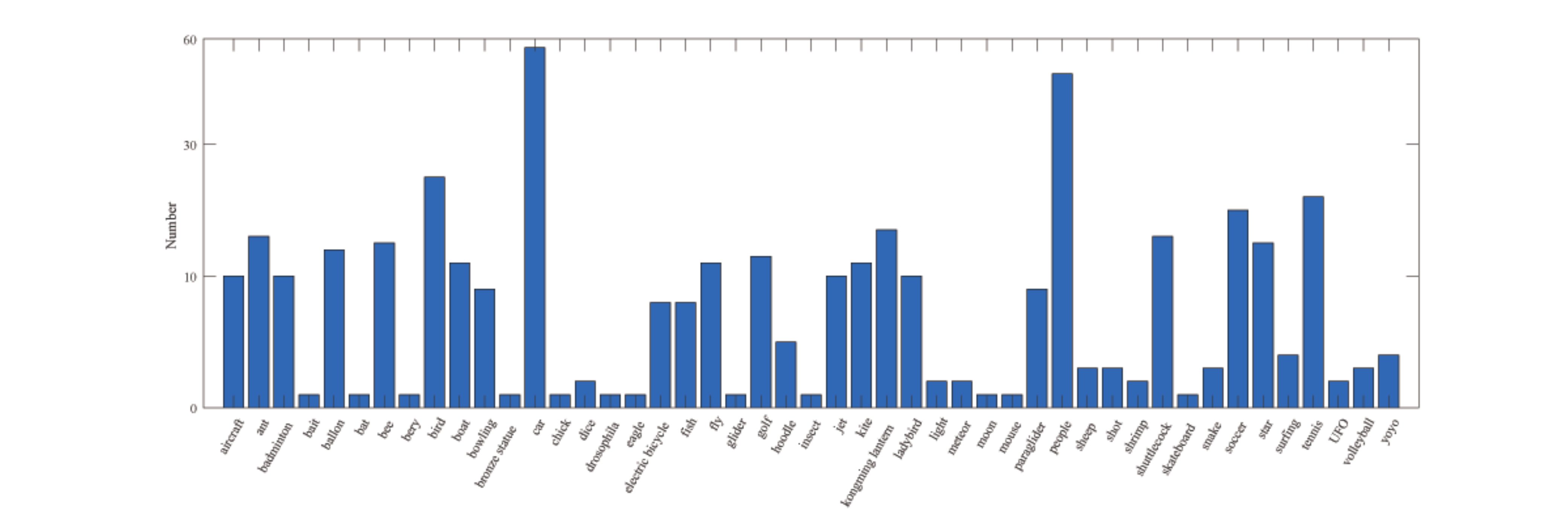} \\
\caption{Data distribution on object classes. }\label{fig::class-number}
\end{figure*}

\subsection{Statistical Analysis}

{\flushleft \bf Object Category}.
The diversity of object categories helps to test the performance of trackers on different categories of tiny objects. In LaTOT, we elaborately collect 48 categories for tiny objects from 270 scenes, and the distribution of video numbers are reported in Fig.~\ref{fig::class-number}.
We notice that the data distribution of LaTOT on object categories conforms to the long-tail distribution, in which the learning under this unbalanced data is an important topic in practical applications.
It can encourage the exploration of more practical and extensible tiny object tracking method.

\begin{table*}[t]\footnotesize 
\setlength{\tabcolsep}{0.6cm}
\caption{Definition of 12 attributes in the proposed LaTOT dataset.} 
\centering 
\renewcommand\arraystretch{1.5}
\begin{tabular}{l|l}
\hline
\bf Attribution & \bf Description \\
\hline
\bf SV & Scale Variation: the ratio of bounding box area is outside the range [0.5, 2] after 1s. \\

\bf FM & Fast Motion: the motion of the ground truth bounding box is larger than the size of the bounding box. \\

\bf OV & Out-of-View: some portion of the target leaves the camera field of view. \\

\bf IV & Illumination Variation: the illumination of the target changes significantly. \\

\bf CM & Camera Motion: abrupt motion of the camera. \\

\bf MB & Motion Blur: the target region is blurred due to the motion of target or camera. \\

\bf BC & Background Clutter: the background near the target has the similar color or texture as the target. \\

\bf SO & Similar Object: there are object of similar shape or same type near the target. \\


\bf PO & Partial Occlusion: the target is partially occluded. \\

\bf FO & Full Occlusion: the target is fully occluded. \\

\bf AM & Abrupt Motion: Abrupt motion of the camera or target. \\


\bf LI & Low Illumination: The illumination intensity in the target area is low.\\\hline
\end{tabular}
\label{tb::att-define}
\end{table*}

\begin{figure}[htb]
\centering
\includegraphics[width=\columnwidth]{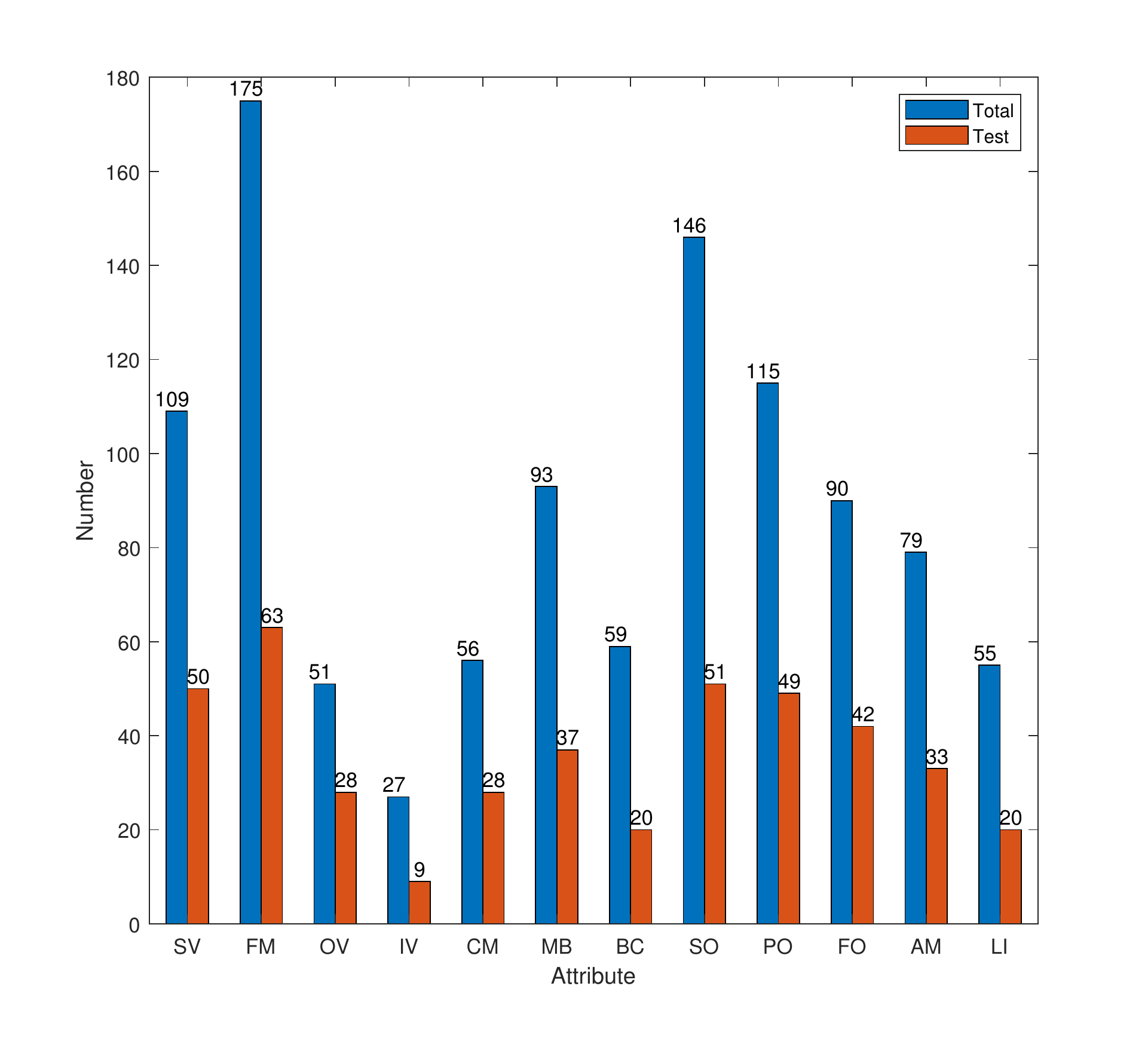} \\
\caption{Distribution of sequences in each attribute on our dataset.}\label{fig::attribute-distribution}
\end{figure}

\begin{figure}[htb]
\centering
\includegraphics[width=\columnwidth]{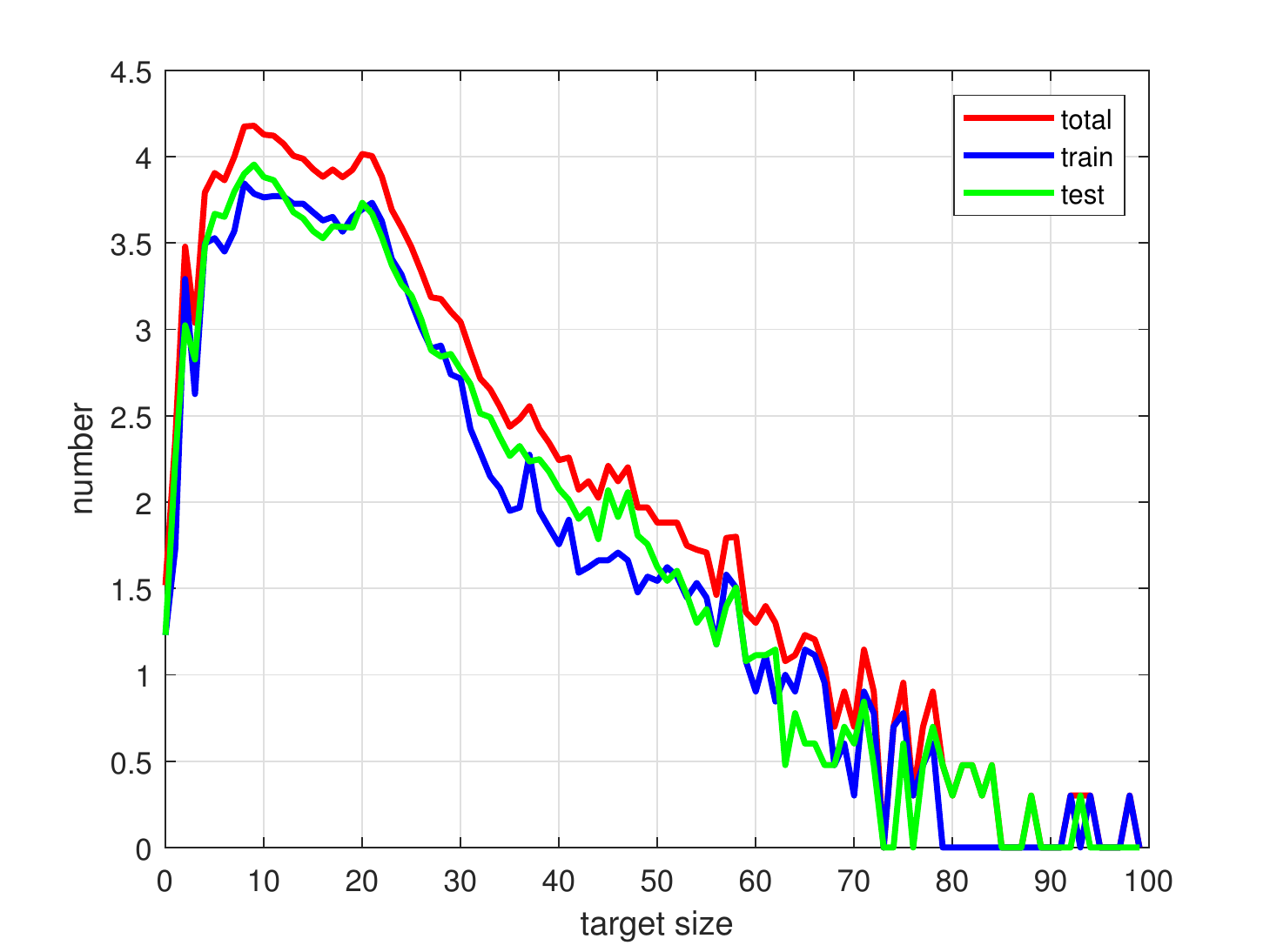} \\
\caption{Scale Distribution. The ordinate is a logarithmic operation of lg.}\label{fig::sd}
\end{figure}

{\flushleft \bf Challenge Attributes}.
To test the performance of different tracking algorithms on various challenge attributes, we define 12 kinds of challenges as shown in Table \ref{tb::att-define}. We also show the distribution of each attribute in our dataset in Fig. \ref{fig::attribute-distribution}. Note that some challenges defined in other tracking datasets are removed, including deformation, in-plane rotation and out-of-plane rotation. Because the tiny objects lack sufficient appearance information to aware these challenges. 


Due to the unique property of tiny objects, trackers are very fragile to track such targets. Here, we give a detailed analysis of the two main factors that affect the performance of tiny object tracking. 
{\bf 1) Low resolution. } Although low resolution attribute are not included in the attribute table, they are presented in all video sequences. Tiny objects with low resolution attribute lack enough appearance information. However, many existing trackers are modelled based on the appearance of object. Therefore, low resolution is a tremendous challenge for these trackers. Moreover, the sizes of tiny objects are very small, and thus directly inputting them into the deep networks will cause their features to disappear. If input image is enlarged directly, it will not only cause image blur and sawtooth phenomenon, which further damage the quality of tiny object image, but also increase the cost of memory and computation. In addition, if a shallow CNN network is used, it is difficult to capture enough target semantic information.
{\bf 2) Background interference. }
Compared with normal object images, tiny objects have small proportions in the whole image, which are easy to be interfered by cluttered and similar background. Especially, by observing PR curves in Fig.~\ref{fig::overall-test-pr-sr}, the overall curves with distance threshold greater than 5 are all relatively flat, which indicates that once the target is lost, it is difficult for trackers to track the target again. It is because the general re-detection based tracking algorithms \cite{voigtlaender2020siam,dai2020high} will expand search range and even carry out full image search to track the target object, once the object is disappears. Due to the particularity of tiny objects, it is easy to be interfered by a large number of background objects and noises. The typical examples can be observed by the full image search algorithm GlobalTrack and the re-detection tracking algorithm Siam R-CNN in Fig. \ref{fig::overall-test-pr-sr}.

{\flushleft \bf Dataset Splitting.}
It is necessary to provide a training set to train deep trackers for tiny object tracking.
To this end, our LaTOT is divided into a training set and a testing set. In specific, 269 of which are divided into training set with 104,910 frames, and the rest video sequences are used as testing set with 112,780 frames. It is worthy noting that we carefully select 260 video sequences with representative challenges in advance, and then randomly selected 165 of them as the testing set.


\section{Multilevel Knowledge Distillation Network}
In this section, we will present the details of the Multilevel Knowledge Distillation Network (MKNet), including the baseline tracker, multilevel knowledge distillations, and the implementation details in training and tracking phases.

\subsection{Baseline Tracker: Super-DiMP}
The baseline approach used in our experiments is the Super-DiMP, which combines the standard DiMP~\cite{bhat2019learning} classifier and PrDiMP~\cite{danelljan2020probabilistic} boundary box regressor. Specifically, DiMP consists of a template branch and a test branch. In the training phase, the template branch is used to learn a target model, and the test branch is utilized to calculate the loss of the task. The target model $f$ is used to distinguish the appearance of foreground and background in the feature space $\mathcal{X}$, and it is actually the filter weights of convolutional layers. The target model $f$ can be optimized by the following loss:
\begin{equation}
L(f)=\frac{1}{\left|S_{\text{train }}\right|} \sum_{(x, c) \in S_{\text{train }}}\|r(x * f, c)\|^{2}+\|\lambda f\|^{2},\label{lf}
\end{equation}
where $x$ and $c$ are the training samples and the corresponding ground truth labels, $S_{\text {train }}=\left\{\left(x_{j}, c_{j}\right)\right\}_{j=1}^{n}$ is the training set, which consists of training samples $x_{j}$ and ground-truth $c_{j}$. $r(s, c)$ is the residual function, which is used to calculate the difference between the target score map $s=x * f$ and the label $c$ at each spatial position.
After computing the target model $f(i)$ based on Eq.~\eqref{lf}, the network is optimized by minimizing the classification loss on test samples:
\begin{equation}
L_{\mathrm{cls}}=\frac{1}{N_{\mathrm{iter}}} \sum_{i=0}^{N_{\mathrm{iter}}} \sum_{(x, c) \in S_{\text {test }}}\left\|\ell\left(x * f^{(i)}, z_{c}\right)\right\|^{2},
\label{cls}
\end{equation}
Herein, the regression label $z_{c}$ is set to a Gaussian function centered as the target $c$. $N_{iter}$ represents the number of optimization iterations, and  $\ell(s, z)$ is a hinge-like residual function. 

\begin{figure*}[htb]
\centering
\includegraphics[width=1 \textwidth]{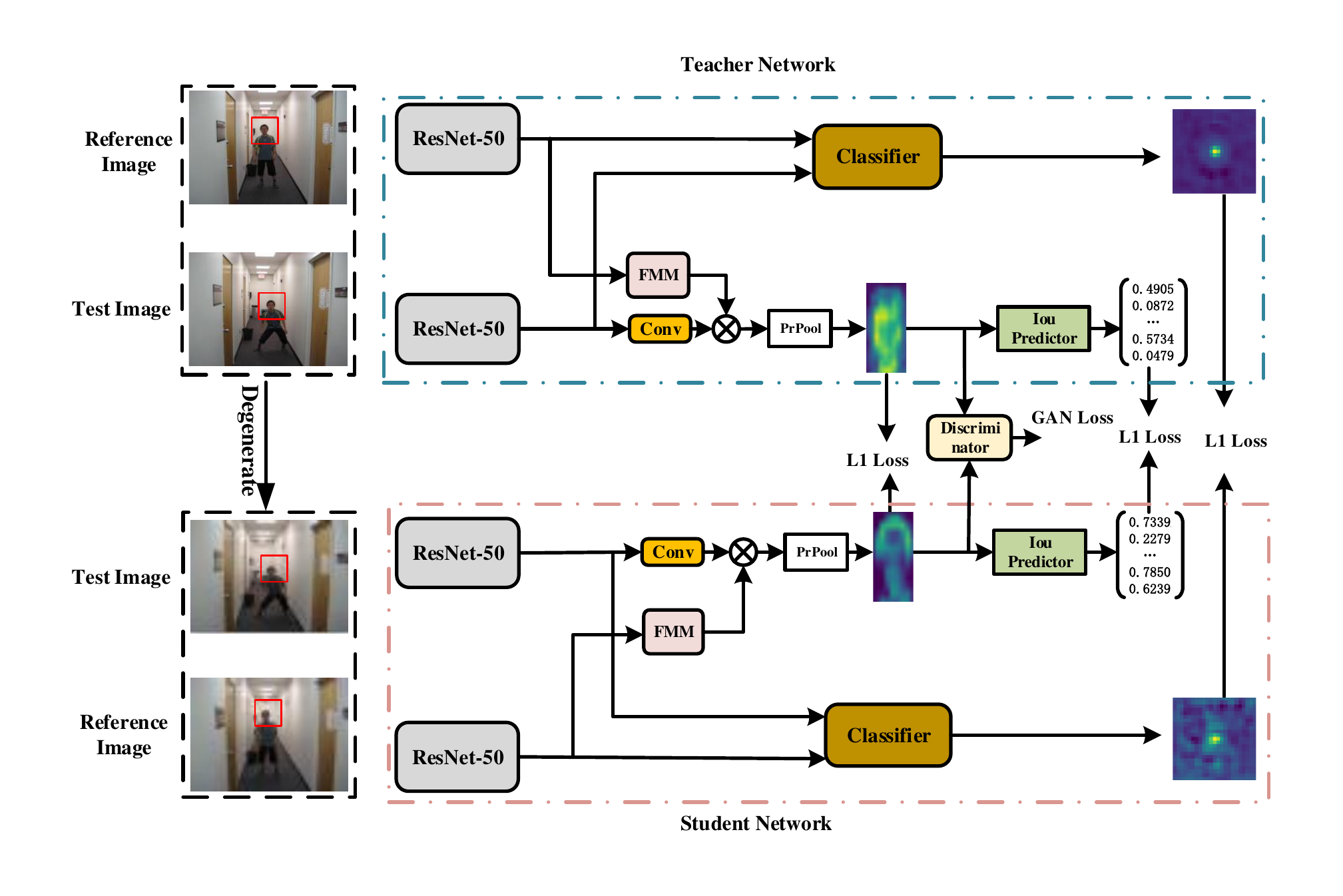} \\
\caption{Framework of MKDNet for tiny object tracking. The `FMM' is a feature modulation module, which contains a $3\times 3$ convolution, a PrPool operation and two $1\times 1$ convolutions.}\label{fig::frame-work}
\end{figure*}

For the IoU regression head network, the baseline network adopts the same structure and loss function with PrDiMP \cite{danelljan2020probabilistic}. Since the ground-truth bounding box is not always accurate, sometimes the center point of the bounding box may not fall on the object, but on the background. 
To eliminate these effects, PrDiMP proposes a probabilistic regression model, which takes the predicted probability distribution $p\left(y \mid x_{i}, \theta\right)$ as the output. This $p\left(y \mid x_{i}, \theta\right)$ can be written as:
\begin{equation}
p(y \mid x, \theta)=\frac{1}{Z_{\theta}(x)} e^{s_{\theta}(y, x)}, Z_{\theta}(x)=\int e^{s_{\theta}(y, x)} \mathrm{d} y\label{p}
\end{equation}
where $x$, $y$ are the input and output of network, $s_{\theta}(y, x)$ represents the prediction confidence value. It is worthy noting that Eq.~\eqref{p} is a general expression of the ${SoftMax} $ operation in continuous space.
The Kullback–Leibler(KL) divergence is used to measure the distance between the predicted density $p\left(y \mid x_{i}, \theta\right)$ and the conditional ground truth distribution $p\left(y \mid y_{i}\right)$ for modeling the label noise and the ambiguity in tracking. The formula of KL divergence is as follows:
\begin{equation}
\begin{array}{l}
\mathrm{KL}\left(p\left(\cdot \mid y_{i}\right), p\left(\cdot \mid x_{i}, \theta\right)\right)=\int p\left(y \mid y_{i}\right) \log \frac{p\left(y \mid y_{i}\right)}{p\left(y \mid x_{i}, \theta\right)} \mathrm{d} y \\
\sim \log \left(\int e^{s_{\theta}\left(y, x_{i}\right)} \mathrm{d} y\right)-\int s_{\theta}\left(y, x_{i}\right) p\left(y \mid y_{i}\right) \mathrm{d} y \label{kl}
\end{array}
\end{equation}
where $\sim$ is an equivalent symbol, which means that a constant term that does not affect the network training is deleted. 
For more details, please refer to  DiMP~\cite{bhat2019learning} and PrDiMP~\cite{danelljan2020probabilistic}.

\subsection{Overview}
As shown in Fig. \ref{fig::frame-work}, our proposed MKDNet mainly contains two sub-networks, i.e., a teacher network and a student network. Note that the two sub-networks share the same network architecture (i.e., the baseline tracker Super-DiMP). The major difference between two networks is their inputs. Specifically, the input of teacher network is the high-resolution object image, while the input of student network is the low-resolution object image degraded from high-resolution one. In the training phase, we use feature-, score-, and IoU-level distillation strategies to train the student network under the guidance of teacher network, which aims at making the student network approaches or even exceeds the teacher network. More importantly, we design a reliable distillation measure to avoid wrong and invalid distillations in every batches, which takes the loss of the teacher network as the distillation upper bound, to control the distillation process. In the testing phase, we only use the student network to carry out tracking.

\subsection{Multilevel Knowledge Distillation}
In this section, we will describe the details of multilevel knowledge distillation network, including image degradation, feature-, score- and IoU-level distillation strategies.

{\flushleft \bf Image Degradation.}
Because existing tracking datasets do not have paired high-resolution and low resolution images. Therefore, we have to degrading images in existing tracking datasets to low-resolution one for the simulation of tiny objects. First, we calculate the average size of ground truth bounding boxes of objects in each batches, and divide it by a scaling factor (we set it as 16 in this work). Then, the bicubic interpolation is used to downsample input images. Since the input image size of the network is fixed, we randomly choose nearest or bilinear interpolation to upsample images to the fixed size, and the input size is $352 \times 352$ in this work.

\begin{figure}[htb]
\centering
\includegraphics[width=\columnwidth]{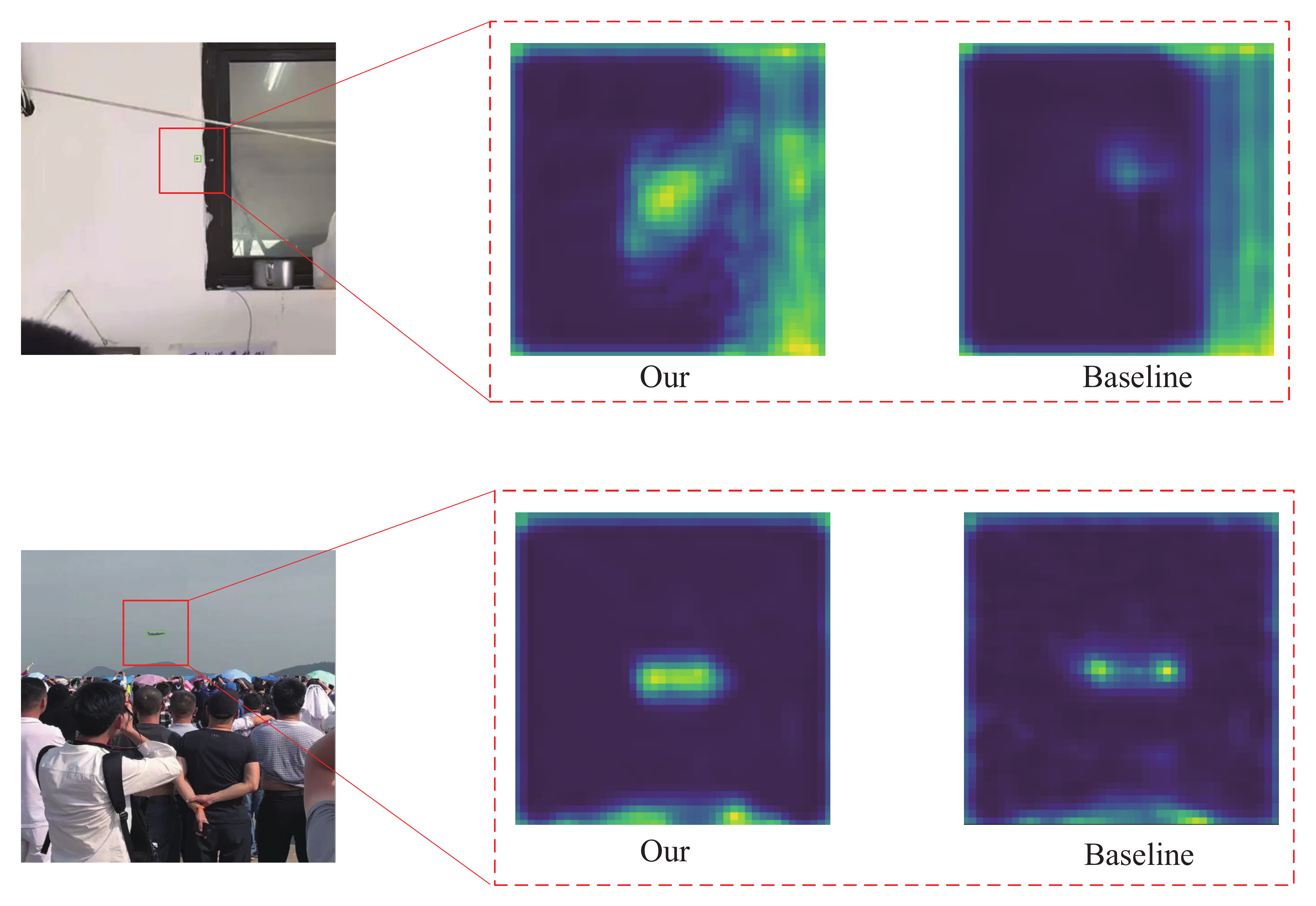} \\
\caption{Illustration of the feature maps of our method against the baseline tracker Super-DiMP.}\label{fig::tip-feature-map}
\end{figure}

{\flushleft \bf Feature-level Distillation.}
Due to the lack of sufficient appearance information, it is difficult to extract high-quality features of tiny objects. To transform low-resolution features into high-resolution ones, we design a feature-level distillation module based on generative adversarial learning. We treat the feature extractor of the network as the generator, and utilize the high-resolution features of teacher network to guide the learning of low resolution features of tiny objects. We design a discriminator which contains three fully connected layers to distinguish the true and false labels of high-resolution features and generated features. The generator and discriminator can be optimized in an alternative manner: 
\begin{equation}
\begin{aligned}
L_{\text {gen} } &=-\sum_{i=1}^{N} \log D\left(\mathbf{S}_{i}^{f}\right) \\
L_{\text {dis} } &=-\sum_{i=1}^{N}\left(\log D\left(\mathbf{T}_{i}^{f}\right)+\log \left(1-D\left(\mathbf{S}_{i}^{f}\right)\right)\right)
\end{aligned}
\end{equation}
where $\mathbf{S}_{i}^{f}$,  $\mathbf{T}_{i}^{f}$ respectively represent the features of student network and teacher network. During this process, we design a feature consistency $\ell_{1}$ loss  as:
\begin{equation}
L_{\text {cons }}=\frac{1}{N}\sum_{i=1}^{N}\left\|\mathbf{S}_{i}^{f} - \mathbf{T}_{i}^{f}\right\|_{1}\label{l_cons}
\end{equation}

To avoid irrelevant background interference, MKDNet uses the features of the region of interest extracted by PrPool~\cite{jiang2018acquisition} for distillation learning, instead of features of the entire image used in existing feature-level knowledge distillation networks~\cite{komodakis2017paying}. As shown in Fig.~\ref{fig::tip-feature-map}, the feature representation ability of our tracker for tiny object is significantly improved against the baseline tracker.

{\flushleft \bf Score-level Distillation.}
Note that the potential distribution of the output prediction of teacher network can be a guidance for the learning of student network~\cite{hinton2015distilling, park2021learning}. Therefore, we design a score-level distillation scheme, which uses the classification loss of teacher as the supervision to teach the learning of student network, so as to improve the discrimination ability of student network. The loss of score-level loss can be expressed as:
\begin{equation}
L_{\text {score-d }}=\frac{1}{N}\sum_{i=1}^{N}\left\|\mathbf{S}_{i}^{score} - \mathbf{T}_{i}^{score}\right\|_{1}\label{l_score}
\end{equation}
where $\mathbf{S}_{i}^{score}$ and $\mathbf{T}_{i}^{score}$ represent the classification score maps of student and teacher networks respectively.

{\flushleft \bf IoU-level Distillation.}
To improve the localization ability of student network, we also design a IoU-level distillation scheme that uses the IoU score of teacher as the supervision to guide the learning of student network. The IoU-level distillation loss can be expressed as:
\begin{equation} 
L_{\text {iou-d }}=\frac{1}{N}\sum_{i=1}^{N}\left\|\mathbf{S}_{i}^{iou} - \mathbf{T}_{i}^{iou}\right\|_{1}\label{l_iou}
\end{equation}
where $\mathbf{S}_{i}^{iou}$ and $\mathbf{T}_{i}^{iou}$ indicate the IoU score of student and teacher networks respectively. The score-level and IoU-level distillation schemes can also be regarded as regularization of MKDNet which can avoid overfitting to some extent.

{\flushleft \bf Reliable Distillation Measure. }
Eq. \eqref{l_cons}, \eqref{l_score} and \eqref{l_iou} actually require that the teacher model is better than the student network in every batch, however, it may not always hold up in the training phase. As the teacher network sometimes has wrong guidance supervision and direct usage of information from teacher network would lead to misleading learning of student network. To address this issue, we design a reliable distillation measure which takes the loss of teacher network as the distillation upper bound, to control the distillation process. Specifically, when the loss of teacher is greater than the loss of student network, the signal from teacher network will be ignored directly. The reliable distillation measure can be written as:
\begin{equation}
\mathrm{RDM}^{iou}=\max (0, \mathcal{L}_{s}^{iou}-\mathcal{L}_{t}^{iou})
\end{equation}
\begin{equation}
\mathrm{RDM}^{cls}=\max (0, \mathcal{L}_{s}^{cls}-\mathcal{L}_{t}^{cls})
\end{equation}
where $\mathrm{RDM}^{iou}$ and $\mathrm{RDM}^{cls}$ are the reliable distillation measure of regression branch of IoU and classification branch respectively.

\subsection{Training and Tracking Phases} 

{\flushleft \bf Offline Training. }
We train the student network on pairs of sets $(M_{\text {train }}, M_{\text {test }})$. Each set $M=\left\{\left(I_{j}, b_{j}\right)\right\}_{j=1}^{N_{\text {frames }}}$ is made up of images $I_{j}$ paired with their corresponding object bounding box labels $b_{j}$. Based on original losses (classification loss (Eq. \eqref{cls}) and IoU loss (Eq. \eqref{kl})), we combine the used distillation loss functions together to train the proposed network:
\begin{equation}
\begin{aligned}
L_{tot}= \alpha L_{\mathrm{cls}}+ \beta L_{i}^{iou}+ \mathrm{RDM}^{iou}*L_{\mathrm{gen}} + \gamma \mathrm{RDM}^{iou}\\ *L_{\text {cons }}
 + \delta \mathrm{RDM}^{cls}*L_{\text {score-d }} + \eta \mathrm{RDM}^{iou}*L_{\text {iou-d }}
\end{aligned}
\end{equation}
where $\mathrm{RDM}^{iou}$ and $\mathrm{RDM}^{cls}$ control whether to carry out distillation and the extent of distillation or not. $\alpha$, $\beta$, $\gamma$, $\delta$ and $\eta$ are set to 100, 0.01, 5, 2 and 0.1 empirically in this work. The first two items are the classification loss and IoU loss, the third and fourth items are the feature-level distillation loss, the fifth item is the score-level distillation loss, and the last item is the IoU-level distillation loss. Following the simultaneous training strategy of GAN, we train $L_{tot}$ and $L_{dis}$ simultaneously. It is worthy noting that we fix the parameters of the teacher network, only train the student network, and load the pre-training model officially provided by Super-DiMP. The entire training process is performed on the training set of TrackingNet \cite{muller2018trackingnet}, LaSOT \cite{fan2019lasot} and GOT-10K \cite{huang2019got}.

{\flushleft \bf Online Tracking. } 
In the online tracking phase, we only use the student network, and the detailed operation of our method is exactly the same as baseline tracker Super-DiMP. Given the first frame with ground truth bounding box, the data augmentation method \cite{bhat2019learning} is used to build an initial training set $S_{train}$ which  including 15 samples. The target model (i.e., convolutional kernel $f$ in Eq. \eqref{lf}) can be learned with discriminative model (i.e., Eq. \eqref{lf}) on $S_{train}$, and is updated every 20 frames or when an interference peak is detected. In addition, the IoU predictor \cite{danelljan2019atom} is used to further refine target bounding boxes.

\begin{figure*}[htb]
\centering
\includegraphics[width=\textwidth]{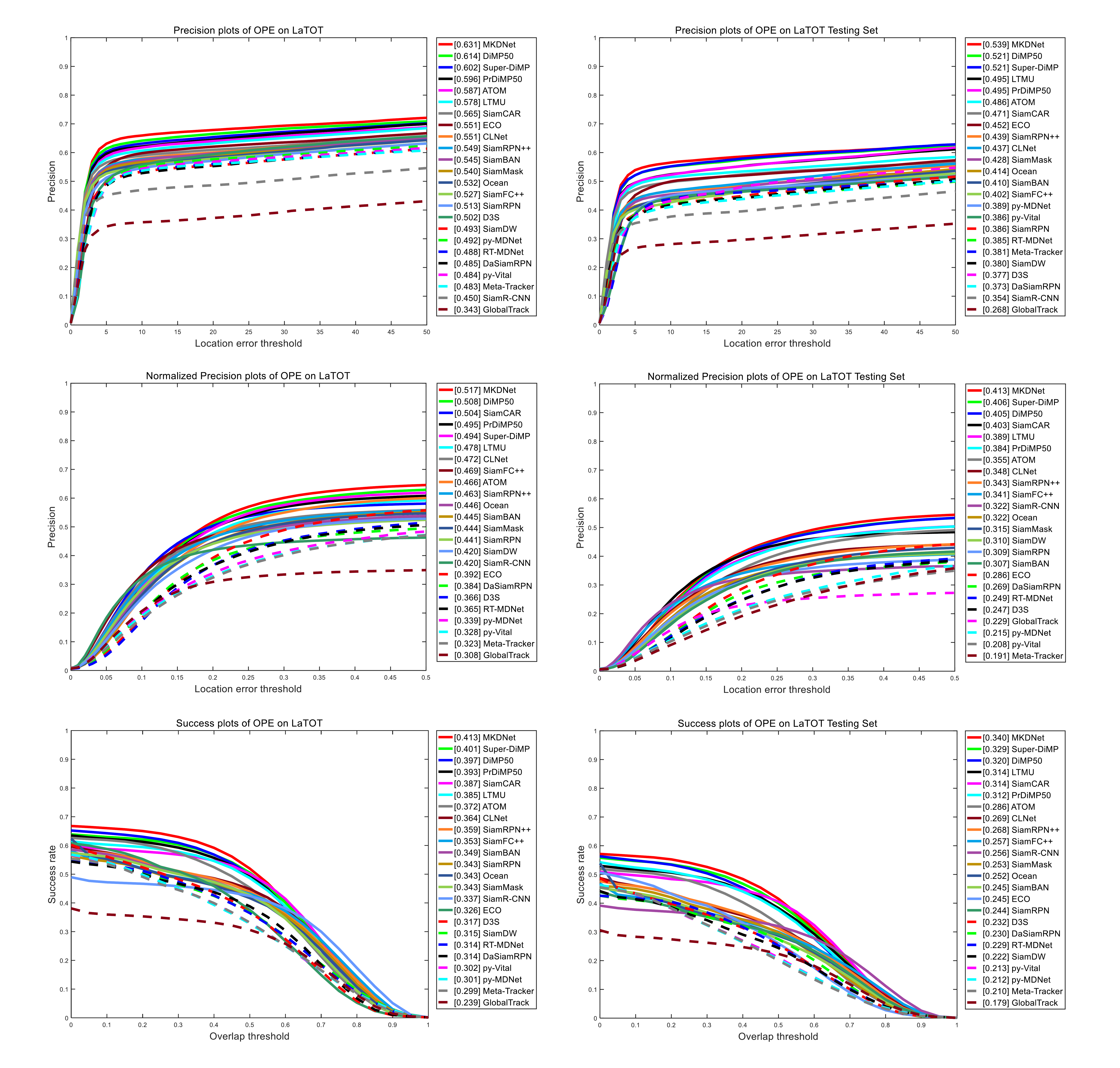} \\
\caption{Performance of trackers on LaTOT and LaTOT testing set with precision plot, normalized precision plot and success plot respectively. The representative scores are shown in the legend. }\label{fig::overall-test-pr-sr}
\end{figure*}

\section{Experiment}

In this section, we conduct extensive experiments on our newly proposed LaTOT benchmark dataset. Specifically, we will first introduce the experimental settings, including implementation details, evaluation protocols, metrics, and baseline trackers. Then, we report our results and compare with other trackers in section \ref{evalResults}. After that, we analyze the tracking results under each challenging attributes in section \ref{attributeResults}. We also give extensive ablation studies, analysis on training data, qualitative evaluation, and failed cases, in the subsequent sub-sections, respectively.

\subsection{Evaluation Setting}
{\flushleft \bf Implementation Details}.
The experiments are conducted on two platforms with I7-9800k CPU (32G RAM), NVIDIA RTX 2080Ti GPU and I9-9900K CPU, NVIDIA RTX 2080Ti GPU. All the experiments of our method is implemented in Python-3.7.9 using PyTorch-1.4 and operates at about 43 FPS with ResNet-50 backbone. We train our network for 50 epochs, and each epoch with 20000 videos. Each video with 3 training and 3 testing images. We employ the ADAM optimizer \cite{kingma2014adam} with learning rate delay of 0.2 every 15 epochs to train the proposed network. The learning rate of classifier and bounding box regressor are $5e-5$ and $5e-4$, respectively. The $layer3$ of backbone trained with the learning rate of $5e-5$.

{\flushleft \bf Evaluation Protocols}.
We use two evaluation protocols for evaluating trackers. 1) All the 434 sequences are adopted to evaluate the performance of tracking algorithms. It aims to provide a large-scale evaluations for tracking algorithms. 2) A testing set is used to evaluate trackers. On one hand, such setting reduces the burden of evaluation of tracking algorithms. On the other hand, it provides a training set for facilitating training deep trackers to handle the challenges of tiny objects.

{\flushleft \bf Evaluated Algorithms}.
We evaluate 24 latest and representative trackers on our benchmark. These tracking algorithms cover the  mainstream researches on current tracking field, i.e., Dimp50~\cite{bhat2019learning}, Super-DiMP, PrDimp50~\cite{danelljan2020probabilistic}, ATOM~\cite{danelljan2019atom}, LTMU~\cite{dai2020high}, SiamCAR~\cite{guo2020siamcar}, SiamRPN++~\cite{li2019siamrpn++}, CLNet~\cite{2020CLNet}, ECO~\cite{danelljan2017eco}, SiamBAN~\cite{chen2020siamese}, SiamMask~\cite{wang2019fast}, SiamFC++~\cite{xu2020siamfc++}, Ocean~\cite{Ocean_2020_ECCV}, SiamRPN~\cite{li2018high}, D3S~\cite{lukezic2020d3s}, SiamDW~\cite{zhang2019deeper}, DaSiamRPN~\cite{zhu2018distractor}, py-MDNet~\cite{nam2016learning}, RT-MDNet~\cite{jung2018real}, py-Vital~\cite{song2018vital}, Meta-Tracker~\cite{park2018meta}, SiamRCNN~\cite{voigtlaender2020siam} and GlobalTrack~\cite{2019GlobalTrack}.

It is worthy to note that the Super-DiMP is an improved version of DiMP, which combines the standard DiMP classifier and PrDiMP boundary box regressor. In addition, the authors of LTMU give a lot of applications of their algorithms on different baselines. We choose the strongest combination on the official homepage of LTMU, namely SuperDiMP+MU.

{\flushleft \bf Evaluation Metrics}.
In this work, we use \textbf{precision rate (PR)}, \textbf{normalized precision rate (NPR)} and \textbf{success rate (SR)} to evaluate all trackers. The precision rate shows the percentage of frames whose estimated locations are below the given distance threshold of the ground truths. To rank the trackers, the threshold is usually set to 20 pixels in \cite{wu2013online}. However, the threshold of 20 is somewhat unreasonable for tiny object tracking, therefore we set the threshold to 5 pixels. As the precision metric is sensitive to target size and image resolution, we normalize the precision over the size of the ground truth bounding box to calculate the normalized precision rate. With the normalized precision metric, we use the Area Under the Curve (AUC) between 0 and 0.5 to rank the tracking algorithm. Success rate is the ratio of successful frames whose overlap of prediction bounding  box and ground truth bounding box is greater than a predefined threshold. In this work, all trackers are ranked by using the AUC between 0 to 1.

\begin{table*}[t]\footnotesize 
\setlength{\tabcolsep}{4mm}
\caption{ The SR scores of the representative tracking methods on different benchmark datasets. $"-"$ means no relevant data.}
\centering
\renewcommand\arraystretch{1.5}
\begin{tabular}{ c | c c c c c c c c}
\hline
Bencrmark & OTB100 & UAV123 & NFS & LaSOT & GOT-10k & TrackingNet & LaTOT(All/Test) \\\hline

LTMU~\cite{dai2020high} & - & - & - & 0.572 & - & - & 0.404/0.334 \\

CLNet & - & 0.633 & 0.543 & 0.499 & - & - & 0.364/0.269 \\

Dimp50~\cite{bhat2019learning} & 0.684 & 0.654 & 0.620 & 0.569 & 0.611 & 0.740 & 0.397/0.320 \\

SiamCAR~\cite{guo2020siamcar} & - & 0.614 & - & 0.507 & 0.569 & - & 0.387/0.314 \\

SiamRPN++ ~\cite{li2019siamrpn++}& 0.696 & 0.613 & 0.502 & 0.496 & 0.517 & 0.733 & 0.359/0.268 \\

SiamR-CNN~\cite{voigtlaender2020siam} & 0.701 & 0.649 & 0.639 & 0.648 & 0.649 & 0.812 & 0.337/0.256 \\

Ocean~\cite{Ocean_2020_ECCV} & 0.684 & - & - & 0.560 & 0.611 & - & 0.343/0.252 \\

ECO~\cite{danelljan2017eco} & 0.700 & 0.537 & 0.466 & 0.324 & 0.316 & 0.554 & 0.326/0.245 \\

RT-MDNet~\cite{jung2018real} & 0.650 & 0.528 & - & 0.325 & 0.404 & - & 0.314/0.229 \\

GlobalTrack~\cite{2019GlobalTrack} & - & - & - & 0.521 & - & 0.704 & 0.239/0.179 \\
\hline
\end{tabular}
\label{tb::datasets-results}
\end{table*}

\subsection{Evaluation Results} \label{evalResults} 
We show the evaluation results of all tracking algorithms on LaTOT and LaTOT testing set in Fig.~\ref{fig::overall-test-pr-sr}, respectively.
From the results we notice that the type of algorithms (DIMP50, Super-Dimp, PrDimp50, ATOM and LTMU) combining Siamese network and IoUNet~\cite{jiang2018acquisition} network have more robust performance. They benefit from good scale regression techniques and online learning strategies. 
SiamCAR is an anchor-free tracker, which performs better than all anchor-based trackers. Compared with other tracking methods, the performance of the anchor-based Siamese tracking algorithms is mediocre. Compared with the trackers with large-scale training, py-MDNet, py-Vital, RT-MDNet and Meta-tracker have poor performance, which might be because the imprecise ridge regressor limits the performance of them. 
GlobalTrack performs searching in whole images, and completely discards the prior information of the target object position in the previous frames. Since the objects in our dataset are too small, it is difficult for the algorithm to effectively extract the features of the small objects, and too much background interference makes it difficult to match the target object. Therefore,  the GlobalTrack performs particularly poorly on our dataset. Looking at Fig.~\ref{fig::overall-test-pr-sr}, we can find that these trackers have poor performance on the testing set, which shows that the testing set is more challenging than the entire LaTOT dataset.
Compared with other algorithms, our method achieves the best results. Specifically,  compared with the baseline tracker and the second best tracker (DiMP-50), MKDNet improves the performance by  2.9\% PR, 2.3\% NPR, 1.2\% SR, and 1.7\% PR, 0.9\% NPR, 1.6\% SR on LaTOT, respectively.  It proves that our multi-level knowledge distillations can effectively alleviate the challenges brought by tiny object.

In addition, we compare the results of some representative algorithms in several mainstream datasets and show them in Table~\ref{tb::datasets-results}. From Table~\ref{tb::datasets-results}, we can clearly see that these algorithms have significantly low accuracy on our dataset. It suggests that our tiny object tracking dataset bring very big challenges for existing tracking algorithms.

\subsection{Attribute-based Results} \label{attributeResults}
To analyze the performance on different challenges, we evaluate 24 trackers on 12 attributes on LaTOT. The results are reported in Table~\ref{tb::attribute}. Compared with other algorithms, our tracker can well mitigate the challenges of background clutter, similar object, illumination variation, partial occlusion, scale variation fast motion camera motion.  It is important to emphasize that these challenges of background clutter, partial occlusion and similar object will significantly affect the quality of object features. These challenges can be handled well by our MKDNet, which proves that our tracker can significantly improve the representation, discrimination and localization abilities for tiny object.

By observing the performance of the challenge attributes of out-of-view,  full occlusion and abrupt motion, we can find that though our method can alleviate some challenge of tiny object tracking to a certain extent, there is still a big gap to solve the problem of tiny object tracking. Therefore, the research and development of tiny object tracking have a long way to go in real-world applications.

\begin{table}[t]\footnotesize
\renewcommand\arraystretch{1.5}
\caption{PR, NPR and SR scores of three versions of our algorithm on LaTOT.}
\centering
\setlength{\tabcolsep}{1mm}{
\begin{tabular}{c |  c    c  c   c c}
	\hline
Metric  &Baseline &MKDNet-FD &MKDNet-SD &MKDNet-IoUD &MKDNet\\\hline
PR     &0.602   &0.616     &0.620     &0.622            &0.631\\
NPR    &0.494   &0.514     &0.513     & 0.513           &0.517\\
SR     &0.401   &0.407     &0.402     & 0.408           &0.413\\\hline
\end{tabular}}
\label{tb::as}
\end{table}

\subsection{Ablation Study}
In this subsection, we will analyze the impact of the main components on tracking performance. In specific, we report the tracking results of three versions of our tracker on LaTOT, as shown in Table~\ref{tb::as}. They are MKDNet-FD, MKDNet-SD and  MKDNet-IoUD. MKDNet-FD, MKDNet-SD and  MKDNet-IoUD respectively indicate that only feature-level, score-level and IoU-level distillation are used. 
From the comparison results in Table~\ref{tb::as}, the PR/NPR score of MKDNet-FD is 1.4\%/2.0\% higher than that of the baseline method, which shows that the proposed feature-level distillation module can effectively enhance the representation ability of tiny object features. Similarly, compared with baseline methods, the PR/NPR score of MKDNet-SD and MKDNet-IoUD  increase 1.8\%/1.9\% and 2.0\%/1.9\% respectively, which shows that score-level distillation and IOU-level distillation can improve the discrimination and localization  ability of tiny object respectively. In addition, compared with other versions, the MKDNet achieves the best results, which shows that the three distillation strategies proposed jointly can achieve better tracking performance.



\begin{table*}[t]\footnotesize 
\setlength{\tabcolsep}{3mm}
\caption{ SR scores of trackers on 12 challenging attributes in LaTOT. Each column of data shows the SR scores of the entire LaTOT dataset. The best and second results are in \textcolor{red}{red} and \textcolor{green}{green} colors, respectively.}
\centering
\renewcommand\arraystretch{1.5}
\begin{tabular}{c |c  c  c  c  c  c  c  c  c  c  c  c  c}
\hline
Tracker & SV  & FM & OV & IV  & CM  & MB &BC &SO &PO & FO & AM & LI \\
\hline
MKDNet  &  \textcolor{red}{0.441} &\textcolor{red}{0.264} & \textcolor{green}{0.171} & \textcolor{red}{0.474} & \textcolor{red}{0.378} & 0.141 & \textcolor{red}{0.404} & \textcolor{red}{0.448} & \textcolor{red}{0.299} & \textcolor{green}{0.227} & 0.156 & \textcolor{green}{0.522} \\

LTMU~\cite{dai2020high} &\textcolor{green}{0.43} & \textcolor{green}{0.261} & 0.169 &\textcolor{green}{0.46} & 0.355 & 0.148 & 0.373 & 0.416 & 0.281 &0.22 &  {0.157} & 0.514 \\

Super-DiMP~\cite{danelljan2020probabilistic}  & 0.424 & 0.257 & 0.164 & 0.445 & \textcolor{green}{0.371} & 0.147 &\textcolor{green}{ 0.386} & 0.422 & \textcolor{green}{0.291} & 0.218 & 0.155 & 0.517   \\

PrDiMP50~\cite{danelljan2020probabilistic} & 0.429 & 0.26 & 0.157 & 0.445 & 0.362 & \textcolor{green}{0.149} & 0.361 & 0.41 & 0.278 & 0.213 & \textcolor{green}{0.161} & 0.471 \\

DiMP50~\cite{bhat2019learning} & 0.407 & 0.256 & 0.141 &\textcolor{red}{0.474} & 0.358 & 0.122 & 0.382 & \textcolor{green}{0.437} & 0.271& 0.205 & 0.138 &\textcolor{red}{0.527}   \\

SiamCAR~\cite{guo2020siamcar}  & 0.377 & 0.209 & 0.119 & 0.424 & 0.313 & 0.102 & \textcolor{green}{0.386} & \textcolor{red}{0.448} & 0.241 & 0.163 & 0.109 & 0.52   \\

ATOM~\cite{danelljan2019atom} & 0.379& 0.226 & 0.132 & 0.429 & 0.333 & 0.0999 & 0.355 & 0.418 & 0.236 & 0.19 & 0.111 & 0.494   \\

CLNet~\cite{2020CLNet}  & 0.384 & 0.209& 0.111 & 0.389 & 0.303 & 0.0905 & 0.338 & 0.391 & 0.229 & 0.141 & 0.105 & \textcolor{red}{0.527}  \\

SiamRPN++~\cite{li2019siamrpn++}   & 0.368 & 0.196 & 0.116 & 0.401 & 0.288 & 0.0828 & 0.339 & 0.393 & 0.239 & 0.145 & 0.0998 & 0.521  \\

SiamFC++~\cite{xu2020siamfc++} & 0.377 & 0.2 & 0.109 & 0.401 & 0.316 & 0.0828 & 0.333 & 0.371 & 0.237 & 0.138 & 0.0953& 0.511   \\

SiamBAN~\cite{chen2020siamese} & 0.373 & 0.212 & 0.111 & 0.42 & 0.286 & 0.0976 & 0.312 & 0.377 & 0.23 & 0.163 & 0.106 & 0.485   \\

SiamMask~\cite{wang2019fast} & 0.367 & 0.196 & 0.111 & 0.383 & 0.286 & 0.0832& 0.322 & 0.375 & 0.218 & 0.137 & 0.0922 & 0.505   \\

SiamRPN~\cite{li2018high} & 0.361 & 0.187 & 0.12 & 0.386 & 0.296 & 0.0843& 0.316 & 0.373 & 0.204 & 0.13 & 0.0875 & 0.495   \\

SiamR-CNN~\cite{voigtlaender2020siam} & 0.405 & 0.242 & \textcolor{red}{0.193} & 0.339 & 0.322 & \textcolor{red}{0.176} & 0.334 & 0.28 & 0.27 & \textcolor{red}{0.228} & \textcolor{red}{0.169} & 0.471  \\

ECO~\cite{danelljan2017eco} & 0.32 & 0.187 & 0.112 & 0.382 & 0.295 & 0.0802 & 0.278 & 0.358 & 0.205 & 0.131 & 0.092 & 0.463   \\

Ocean~\cite{Ocean_2020_ECCV}  & 0.357 & 0.209& 0.125 & 0.395 & 0.272 & 0.0936& 0.3 & 0.363 & 0.208 & 0.142 & 0.0968 & 0.498   \\

DaSiamRPN~\cite{zhu2018distractor} & 0.303 & 0.172 & 0.106 & 0.391 & 0.262 & 0.0683 & 0.277 & 0.356& 0.193 & 0.105 & 0.083 & 0.454   \\

SiamDW~\cite{zhang2019deeper} & 0.3 & 0.167 & 0.102& 0.379 & 0.282 & 0.0879 & 0.26 & 0.349 & 0.199 & 0.108 & 0.0864 & 0.49   \\

D3S~\cite{lukezic2020d3s}  & 0.339 & 0.205 & 0.122 & 0.375 & 0.308 & 0.102 & 0.277 & 0.332 & 0.216 & 0.142 & 0.101 & 0.436   \\

RT-MDNet~\cite{jung2018real} & 0.282 & 0.154 & 0.0868 & 0.4 & 0.27 & 0.0769 & 0.279 & 0.361 & 0.18 & 0.107 & 0.0774 & 0.468   \\

py-MDNet~\cite{nam2016learning}  & 0.277 & 0.152 & 0.0905 & 0.361 & 0.27 & 0.0751 & 0.286 & 0.342 & 0.2 & 0.096 & 0.0836 & 0.46   \\

py-VITAL~\cite{song2018vital} & 0.276 & 0.151 & 0.093 & 0.354 & 0.258 & 0.0779 & 0.294 & 0.338 & 0.193 & 0.106 & 0.0875 & 0.464   \\

Meta-tracker~\cite{park2018meta} & 0.272& 0.148 & 0.0893 & 0.385 & 0.256 & 0.0666 & 0.283 & 0.344 & 0.184 & 0.106 & 0.0766 & 0.464   \\

GlobalTrack~\cite{2019GlobalTrack}& 0.285 & 0.188 & 0.126 & 0.242 & 0.212 & 0.122 & 0.238 & 0.179 & 0.204 & 0.171 & 0.116 & 0.291   \\
\hline

\end{tabular}
\label{tb::attribute}
\end{table*}

\begin{table}[t]\footnotesize
\renewcommand\arraystretch{1.5}
\caption{Fine-tuned results for ATOM and Super-DiMP on LaTOT testing set.}
\centering
\setlength{\tabcolsep}{0.5mm}{
\begin{tabular}{c| c c c| c c c}
	\hline
Trackers    &&ATOM&  && Super-DiMP& \\\hline
Metric    & PR &NPR &SR & PR &NPR &SR \\\hline
Original training set &0.486 &0.355 &0.286 &0.521 &0.406 &0.329\\
Finetune on LaTOT training set & \bf 0.520 &\bf 0.385 &\bf 0.317 &\bf 0.565 &\bf 0.439 &\bf 0.354 \\\hline

\end{tabular}}
\label{tb::Finetune}
\end{table}

\begin{table}[t]\footnotesize
\renewcommand\arraystretch{1.5}
\caption{Retrained results for Super-DiMP on GOT-10k testing set.}
\centering
\setlength{\tabcolsep}{0.8mm}{
\begin{tabular}{c| c c c| c c c}
	\hline
Trackers    &&ATOM&  && Super-DiMP& \\\hline
Metric    & AO &SR$_{0.50}$ &SR$_{0.75}$ & AO &SR$_{0.5}$ &SR$_{0.75}$ \\\hline
Original training set &0.556 &0.634 &0.402 &0.674 &0.791 &0.59\\
Lasot+GOT-10k+LaTOT & \bf 0.574 &\bf 0.67 &\bf 0.429 &\bf 0.685 &\bf 0.803 &\bf 0.601 \\\hline

\end{tabular}}
\label{tb::Finetune-GOT-10k}
\end{table}

\subsection{Impact of Training Data}
\noindent{\bf Impact on LaTOT testing set.} To show the effects of our training set on the performance of trackers, we load the training model of the trackers and retrain two trackers (Super-DiMP and ATOM). Table~\ref{tb::Finetune} reports the results of Super-DiMP and ATOM on LaTOT testing set and comparisons with the performance of original trackers trained on original tracking datasets. It should be noted that we have not change any hyper-parameters of two trackers. From Table~\ref{tb::Finetune}, we observe that the two trackers gain consistent performance gains, which show the importance of the training set for tiny object tracking.

\noindent{\bf Impact on GOT-10k testing set.}
We retrain the Super-DiMP on LaTOT and the training set of LaSOT and GOT-10K by loading the pre-trained model of Super-DiMP.  For the sake of fairness, all training and inference parameters are kept the same in the experiments. The Table~\ref{tb::Finetune-GOT-10k} shows the tracking results on GOT-10K testing set. Compared with the performance of original Super-DiMP trained on the training set of LaSOT , GOT-10k, TrackingNet and MS COCO~\cite{lin2014microsoft}, our results are significantly improved. It proves that our dataset can effectively boost the performance of deep trackers.

\begin{figure*}[htb]
\centering
\includegraphics[width=\textwidth]{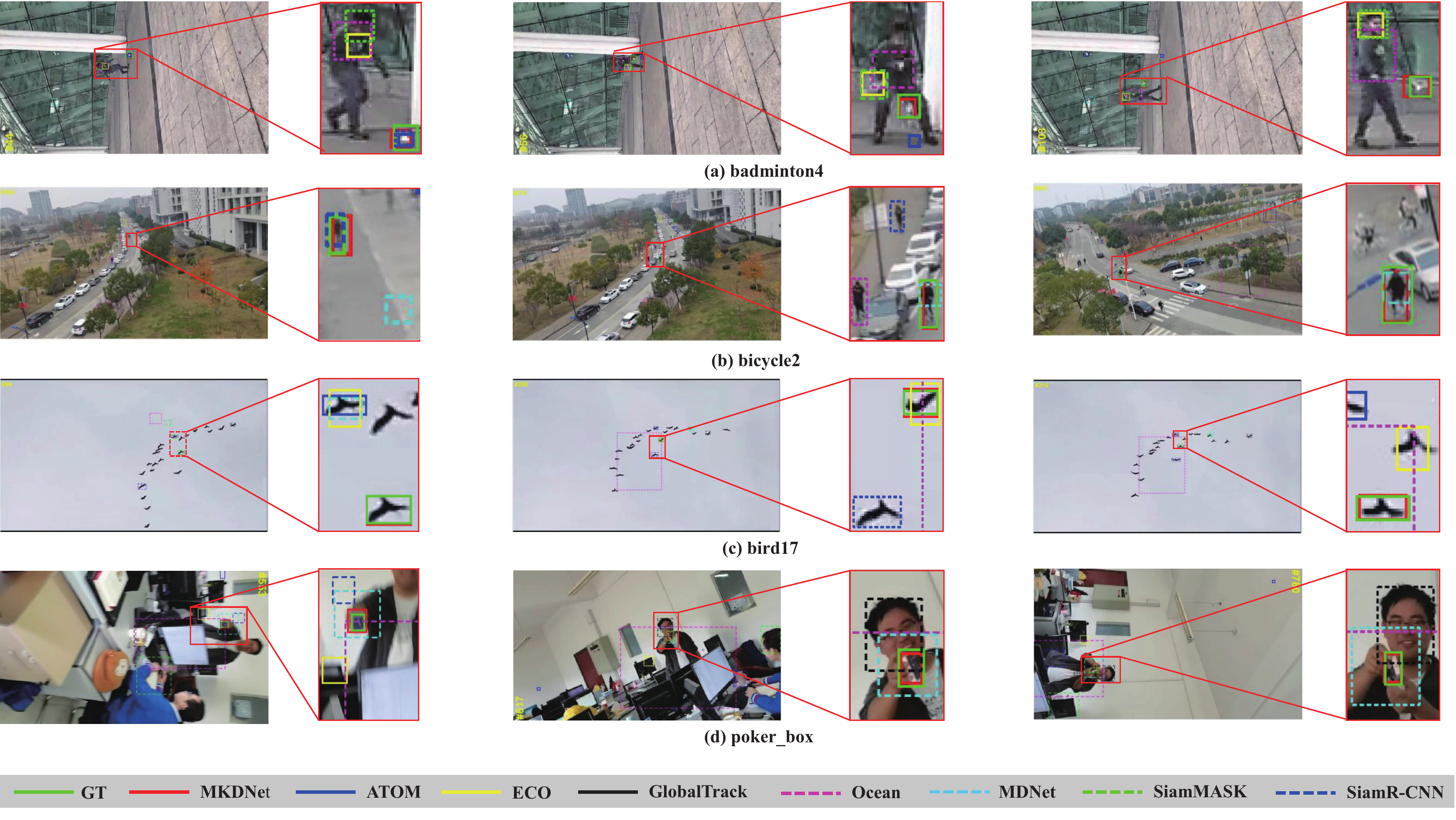} \\
\caption{ Qualitative evaluation on eight representative trackers. To facilitate observation, we enlarge object regions and display it on the right of original images.}\label{fig::Qualitative2}
\end{figure*}

\subsection{Qualitative Evaluation}
To intuitively observe the advantages of our tracker compared to other tracking methods, we qualitatively evaluate seven representative trackers, including ATOM, Ocean, SiamRCNN, ECO, MDNet, SiamMask and GlobalTrack. In Fig.~\ref{fig::Qualitative2}, we show four tracking scenarios with eight challenge attributes, which include low resolution, fast motion, out-of-view, motion blur, similar object, full occlusion, scale variation and abrupt motion. Moreover, to facilitate observation, we enlarge the target object regions and display them on the right of original image.
On LaTOT, multiple challenge attributes often appear in same sequences, which bring a huge challenge for tracking tiny objects and easily lead to failures of current trackers.
For example, as shown in Fig.~\ref{fig::Qualitative2} (a), the sequence of \emph{bidminton4} has the challenges of fast motion, motion blur, out-of-view  and low resolution, making it difficult for current trackers. Thanks to our algorithm, it can significantly improve the feature representation, discrimination and localization abilities for the tiny object to solve these complex challenges compared with other advanced trackers.

 \begin{figure*}[htb]
\centering
\includegraphics[width=\textwidth]{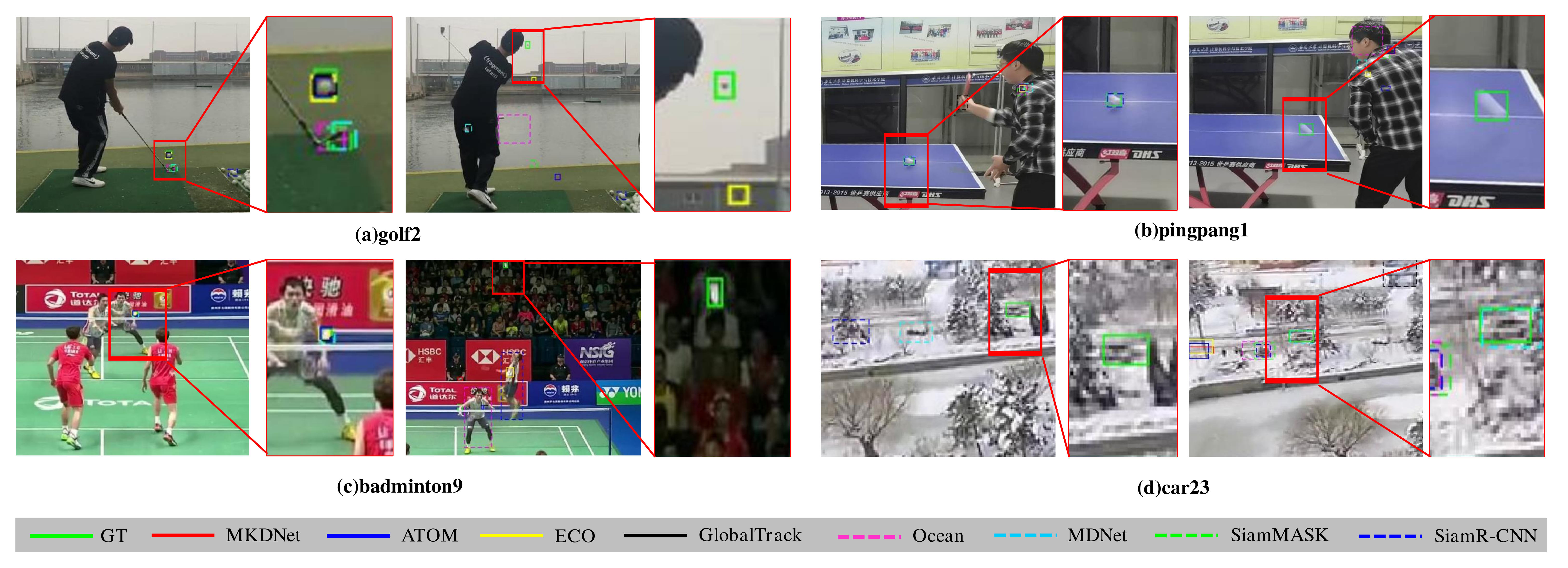} \\
\caption{Failed cases of our tracker and other state of the art trackers.}\label{fig::failed-cases}
\end{figure*}

\subsection{Failed Cases }
As shown in Fig.\ref{fig::failed-cases}, we show some failed cases of our tracker and other state of the art trackers on LaTOT. In these failed cases, there are six challenge attributes, including abrupt motion(as shown in Fig.~\ref{fig::failed-cases}(a) ), motion blur(as shown in Fig.~\ref{fig::failed-cases}(b), (c)), out-of-view(as shown in Fig.~\ref{fig::failed-cases}(c) ), full occlusion(as shown in Fig.~\ref{fig::failed-cases}(d) ), fast motion(as shown in Fig.~\ref{fig::failed-cases}(a), (b), (c)) and background clutter(as shown in Fig.~\ref{fig::failed-cases}(c), (d) ). The challenging problems  of abrupt motion, fast motion and out-of-view will cause target to exceed the search area of the trackers. Even if there are some re-detection tracking algorithms that can solve this kind of problem, because the tiny object lacks enough appearance information and the background interference is too large, these trackers cannot track the object again. Motion blur is also a big challenging problem in LaTOT, which is often accompanied by the challenging attributes of fast motion and camera motion. It will directly affect the representation quality of features. As show in Fig.~\ref{fig::failed-cases}(d), the challenge attributes of full occlusion can easily lead to model drift and targets beyond the search area. To sum up, the main reasons why our tracker and other trackers fail to track are summarized as follows: 1) tiny object have low resolution, more noise, and less effective information, which make trackers unable to effectively extract its features and accurately locate the target. 2) On LaTOT, there are often multiple challenge attributes in same video sequence, which brings a huge challenge to tracking methods.

\section{Conclusion and Future Works}
In this paper, we propose a multilevel knowledge distillation network to effectively enhance the feature representation, discrimination and localization abilities for tiny objects in visual tracking. This method includes three levels of feature-level distillation, score-level distillation, IoU-level distillation to significantly boost the performance of tiny object tracking. To avoid wrong and invalid distillation, a reliable distillation measure is also introduced to control the distillation process. To provide a comprehensive evaluation platform, we present a unified benchmark with high-quality dense annotations, high diversity and challenges for large-scale tiny object tracking. Moreover, we set 12 challenge attributes to evaluate and analyze trackers. Extensive experiments are performed on the proposed dataset, and the results prove the superiority and effectiveness of MKDNet compared with state-of-the-art tracking methods.
The results also show that current trackers still have a large research space in tiny object tracking.

In the future, some potential directions for tiny object tracking can be considered.
We find from Table~\ref{tb::attribute} that the challenges of fast motion, partial occlusion, full occlusion, and motion blur are greatly damaging to the performance of tiny object tracking, which shows that it is difficult to estimate positions of tiny objects directly through limited appearance information.
Therefore, we can employ some trajectory prediction algorithms~\cite{sun2020recursive} to estimate locations of tiny objects by combining the historical location information and appearance information.
Taking the fast motion challenge as an example, trajectory prediction can help the tracker determine the range of target searching area, so as to help accurately locating the target.
In addition, optical flow methods~\cite{ilg2017flownet} can also be explored to predict target appearance states and locations in future frames, thereby help solving the problems of motion blur and fast motion.

\bibliographystyle{IEEEtran}
\bibliography{sample-base}


%





\ifCLASSOPTIONcaptionsoff
  \newpage
\fi

\end{document}